\documentclass[11pt]{article}
\usepackage{naaclhlt2016}
\usepackage{times}
\usepackage{url}
\usepackage{latexsym}
\usepackage{color}
\usepackage{tabularx}
\usepackage{ragged2e}
\newcolumntype{Y}{>{\RaggedRight\arraybackslash}X} 
\usepackage{multirow}
\usepackage{soul}
\usepackage{amsmath}

\usepackage{graphicx}
\usepackage{mathtools}

\naaclfinalcopy

\newcommand{\squishlist}{
 \begin{list}{$\bullet$}
  { \setlength{\itemsep}{0pt}
     \setlength{\parsep}{3pt}
     \setlength{\topsep}{3pt}
     \setlength{\partopsep}{0pt}
     \setlength{\leftmargin}{1.5em}
     \setlength{\labelwidth}{1em}
     \setlength{\labelsep}{0.5em} } }

\newcounter{Lcount}
\newcommand{\squishlisttwo}{
\begin{list}{\arabic{Lcount}. }
{ \usecounter{Lcount}
\setlength{\itemsep}{0pt}
\setlength{\parsep}{0pt}
\setlength{\topsep}{0pt}
\setlength{\partopsep}{0pt}
\setlength{\leftmargin}{2em}
\setlength{\labelwidth}{1.5em}
\setlength{\labelsep}{0.5em} } }

\newcommand{\squishend}{
\end{list} }

\title{Weak Semi-Markov CRFs for NP Chunking in Informal Text}

\author{Aldrian Obaja Muis \and Wei Lu \\
  Singapore University of Technology and Design \\
  {\tt \{aldrian\_muis,luwei\}@sutd.edu.sg}}
  
\date{}

\begin{document}
\maketitle
\begin{abstract}
This paper introduces a new annotated corpus based on an existing informal text corpus: the NUS SMS Corpus \cite{Chen2013}. The  new corpus includes 76,490 noun phrases from 26,500 SMS messages, annotated by university students. We then explored several graphical models, including a novel variant of the semi-Markov conditional random fields (semi-CRF) for the task of noun phrase chunking.
We demonstrated through empirical evaluations on the new dataset that the new variant yielded similar accuracy but ran in significantly lower running time compared to the conventional semi-CRF.
\end{abstract}

\section{Introduction}
 
Processing user-generated text data is getting more popular recently as a way to gather information, such as collecting facts about certain events \cite{Ritter2015}, gathering and identifying user profiles \cite{Layton2010,Li2014,Spitters2015}, or extracting information in open domain \cite{Ritter2012,Mitchell2015a}.

Most recent work focus on the texts generated through Twitter, which, due to the design of Twitter, contain a lot of announcement-like messages mostly intended for general public. In contrast, SMS was designed as a way to communicate short personal messages to a known person, and hence SMS messages tend to be more conversational and more informal compared to tweets.



As conversational texts, SMS data often contains references to named entities such as people and locations relevant to certain events.
Recognizing those references will be useful for further NLP tasks. One way to recognize those named entities is to first create a list of candidates, which can be further filtered to get the desired named entities. Nadeau \cite{nadeau2007survey} lists several methods that work upon candidates for NER. As all named entities are nouns, recognizing noun phrases (NP) is therefore a task that can be potentially useful for further steps in the NLP pipeline to build upon.
Figure \ref{sms-sample-1} shows an example SMS message within which noun phrases are highlighted.
As can be seen from this example, recognizing the NP information on such a dataset presents some additional challenges over conventional NP recognition tasks.
Specifically, the texts are highly informal and noisy, with misspelling errors and without grammatical structures. 
The correct casing and  punctuation information is often missing. 
The lack of spaces between adjacent words makes the detection of NP boundaries more challenging.

Furthermore, the lack of available annotated data for such informal datasets prevents researchers from 
understanding what effective models can be used to resolve the above issues.
In this work, we focus on tackling these issues while making the following two main contributions:

\begin{figure}[t]
\begin{center}
\begin{tabularx}{0.9\columnwidth}{|Y|}
\textit{Hmm \ul{Dr teh} says \ul{the research presentation} should still prepare, but\ul{she}'s not to sure whether \ul{they}'d \ul{time} to present}
\vspace{2.5mm}
\end{tabularx}
\caption{Sample SMS, with NPs underlined}
\label{sms-sample-1}
\vspace{-7mm}
\end{center}
\end{figure}



\squishlist
\item We build a new corpus of SMS data that is fully annotated with noun phrase information.

\item We propose and build a new variant of semi-Markov CRF \cite{Sarawagi2004} for the task of NP chunking on our corpus, which is faster and yields a performance similar to the conventional semi-Markov CRF models.
\squishend


\section{NP-annotated SMS Corpus}
\label{sec-data}

Our text corpus comes from the NUS SMS Corpus \cite{Chen2013}, containing 55,835 SMS messages from university students, mostly in English. 
We used the 2011 version of the corpus, containing 45,718 messages, as it is more relevant to modern phone models using full keyboard layout.


We note that there are a small portion of the messages written in non-English language, such as Tamil and Chinese. 
As we are focusing on English, we excluded messages written by non-native English speakers based on the metadata (21.3\% of all messages). 
We also excluded messages which contain only one word (6.1\%) and we remove duplicate messages (8.1\%). \footnote{We also manually excluded some messages (ID 1017-4016) which are mostly not written in English (4.0\% of all messages).}

We assigned the remaining 27,700 messages to 64 university students who conduct annotations, each annotating 500 with 100 messages co-annotated by two other annotators.
After manual verification we excluded annotations with low quality from 3 students.
We used the resulting 26,500 messages as our dataset.
The students were asked to annotate the top-level noun phrases found in each message using the BRAT rapid annotation tool\footnote{\texttt{http://brat.nlplab.org}}, where they were instructed to highlight character spans to be marked as noun phrases. 
The number of noun phrases per message can be found in Table \ref{sms-np-stat}.

Due to the noisy nature of SMS messages, there may not be proper capitalization or punctuation, and in some cases there might be missing spaces between words. Figure \ref{sms-sample-1} shows a sample SMS message taken from the corpus. We can see that ``Dr teh'' is not properly capitalized and ``she'' in ``butshe's'' is missing spaces around it. NPs which do not have clear boundaries (\textit{improper} NPs) constitutes 4.0\% of all NPs.

We then use this dataset to evaluate some models on base NP chunking task, where, given a text, the system should return a list of character spans denoting the noun phrases found in the text.

\section{Models}
\label{sec-model}


In this paper, we will build our models based on a class of discriminative graphical models, namely conditional random fields (CRFs) \cite{Lafferty2001}, for extracting NPs. The edges in the graph represents the dependencies between states and the features are defined over each edge in the graph. 
Though CRFs are undirected graphical models, we can use directed acyclic graphs with a root, a leaf, and some inner nodes to represent label sequences\footnote{Extension to directed hypergraphs is possible. See \cite{lu-roth:2015:EMNLP}.}.
A path in the graph from the root to the leaf represents one possible label assignment to the input. In the labeled instance, there will be only one single path from the root to the leaf, while for the unlabeled instance, the graph will compactly encode all possible label assignments.
The learning procedure is essentially the process that tries to tune the feature weights such that the true structures get assigned higher weights as compared to all other alternative structures in the graph.

\begin{table}
\small
\begin{center}
\begin{tabular}{r@{\hskip 3pt}|@{\hskip 3pt}r@{\hskip 6pt}r@{\hskip 3pt}|@{\hskip 3pt}r@{\hskip 3pt}|r}
 & \#SMS & \#NPs & \#\textit{improper} & \#tokens \\\hline
total & 26,500 & 76,490 & 3,066 (4.0) & 359,009 \\
train & 21,200 & 61,212 & 2,406 (3.9) & 287,590 \\
dev  & 2,650   & 7,617   & 338 (4.4)    & 35,470 \\
test  & 2,650   & 7,661   & 322 (4.2)    & 35,949 \\
\end{tabular}
\end{center}
\caption{Number of messages, NPs, number of \textit{improper} NPs (as percentage in brackets), which are NPs that do not have clear boundaries, and number of tokens.}
\vspace{-5mm}
\label{sms-np-stat}
\end{table}

In general, a CRF tries to maximize the following objective function:
\begin{eqnarray}
\mathcal{L}(\mathcal{T}) = \hspace*{0.75\columnwidth}\nonumber\\
\sum_{(\mathbf{x},\mathbf{y})\in\mathcal{T}}\!\!\left[\sum_{e\in \mathcal{E}(\mathbf{x},\mathbf{y})} \!\!\!\!\mathbf{w}^T\mathbf{f}(e)
- \log Z_{\mathbf{w}}(\mathbf{x})\vphantom{\sum_{e\in\mathcal{E}(\mathbf{x}}}\right] 
\!\!- \lambda ||\mathbf{w}||^2 \label{eqn-crf}
\end{eqnarray}
where $\mathcal{T}$ is the training set, $(\mathbf{x}, \mathbf{y})$ is a training instance consisting of the sentence $\mathbf{x}$ and the label sequence $\mathbf{y}\in\mathcal{Y}^n$ for a label set $\mathcal{Y}$, $\mathbf{w}$ is the feature weight vector, {\color{black}$\mathcal{E}(\mathbf{x},\mathbf{y})$ is the set of edges which defines the input $\mathbf{x}$ labeled with the label sequence $\mathbf{y}$}, $\mathbf{f}(e)$ is the feature vector of the edge $e$, $Z_{\mathbf{w}}(\mathbf{x})$ is the normalization term which sums over all possible paths from the root to the leaf node, and $\lambda$ is the regularization parameter.

The set of edges and features defined in each model affects the feature expectation and the normalization term. Computation of the normalization term, being the highest in time complexity, will determine the overall complexity of training the model. The set of edges and the normalization term in each model will be described in the following sections.

\begin{figure*}[ht!]
\vspace{-16mm}
\centering
\scalebox{1.0}
{
\def\arraystretch{0.75}\tabcolsep=-1pt
\begin{tabular}{ccc}
\raisebox{0.0mm}{
\includegraphics[scale=0.22]{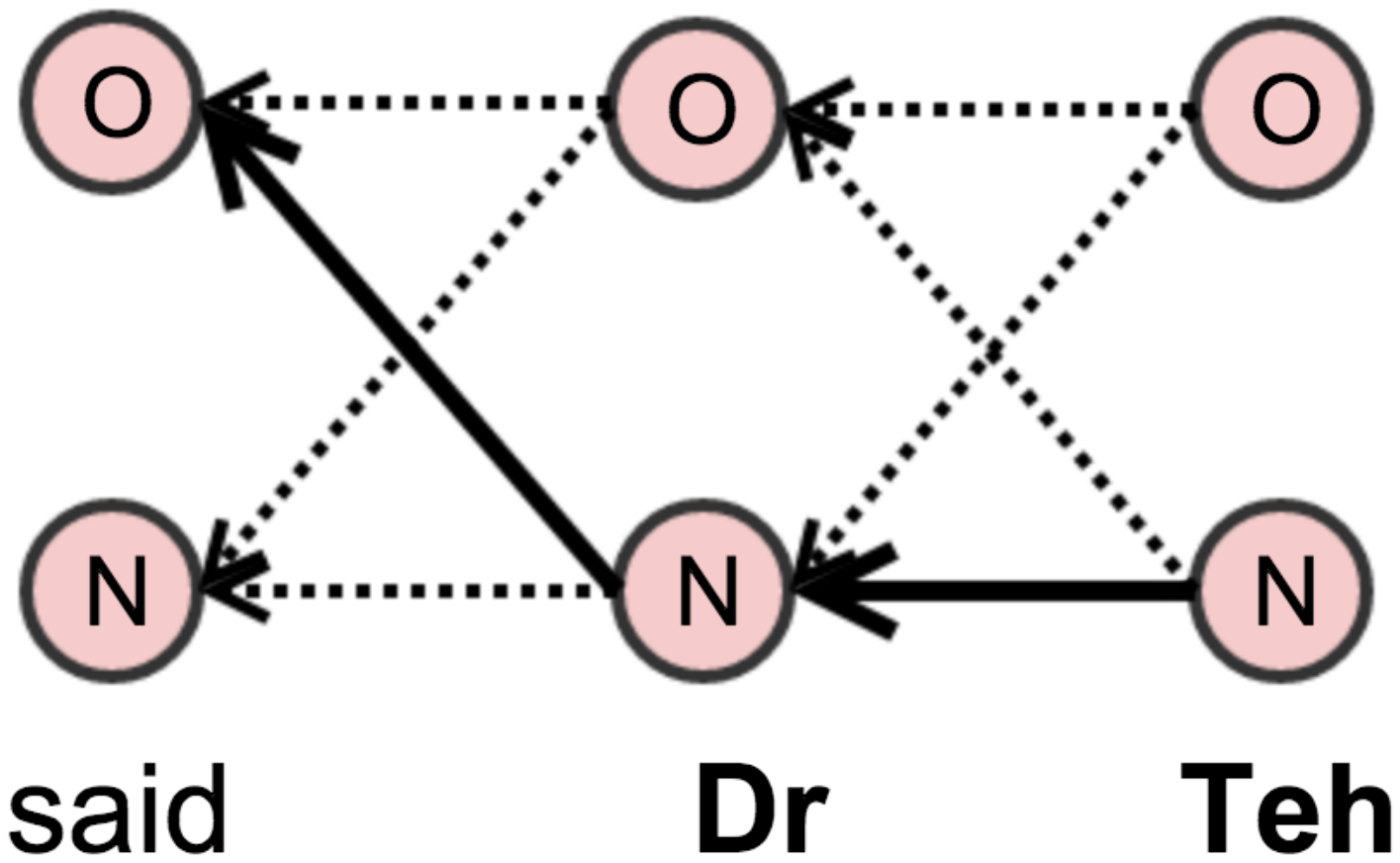}
}
\vspace{-10mm}
&
\raisebox{0.1mm}{
\includegraphics[scale=0.22]{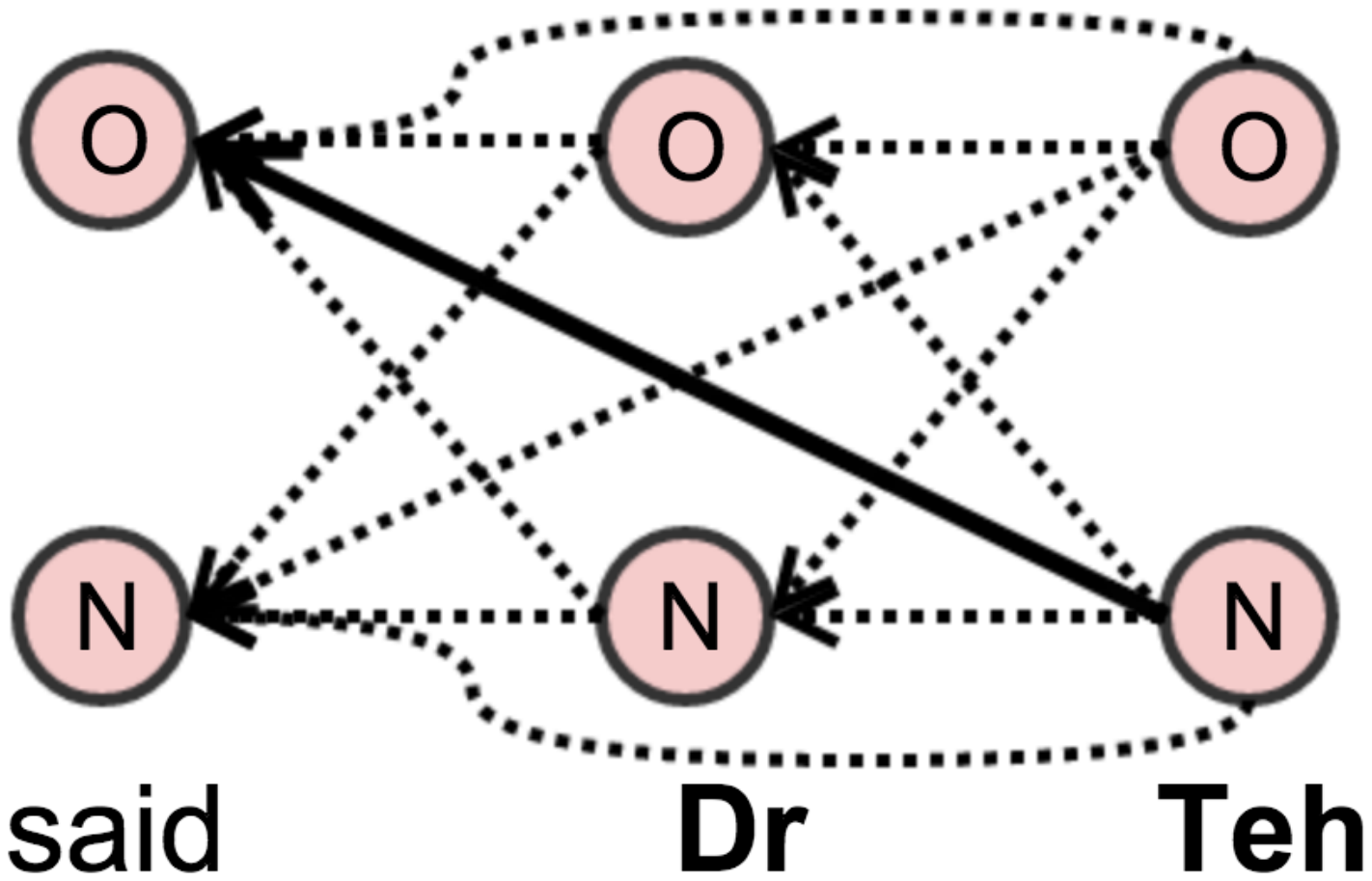}
}
&
\raisebox{11mm}{
\includegraphics[scale=0.22]{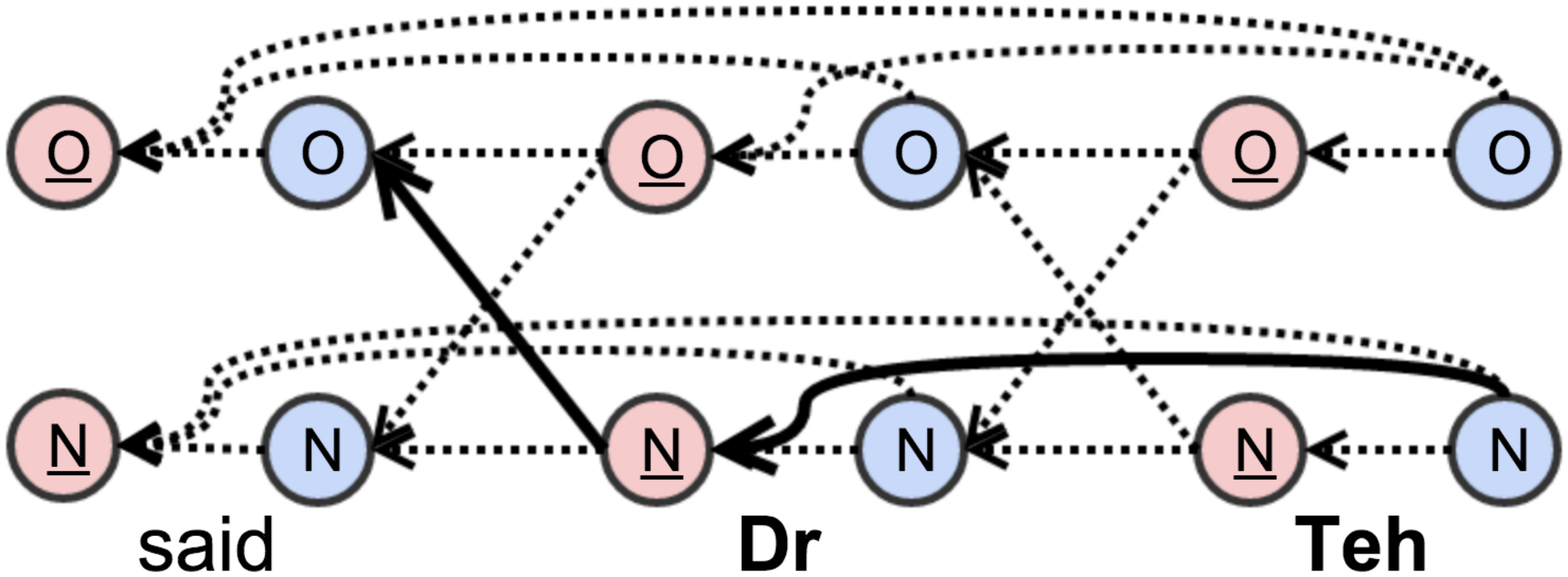}
}
\vspace{-12mm}
\\
Linear CRF&Semi CRF&Weak Semi-CRF\\
\end{tabular}
}
\caption{Graphical illustrations of the differences between three models. The bold arrows represent the path in each model to label ``Dr Teh'' as a noun phrase. For Linear CRF, this is a simplified diagram; in the implementation we used the ``BIO'' approach to represent text chunks. The underlined nodes in Weak Semi-CRF are the Begin nodes.}
\label{fig:diffs}
\end{figure*}

\subsection{Linear CRF}

A linear-chain CRF, or linear CRF is a standard version of CRF which was introduced in \newcite{Lafferty2001}, where each word in the sentence is given a set of nodes representing the possible labels, and edges are present between any two nodes from adjacent words, forming a trellis graph. Here we consider only the first-order linear CRF.

The normalization term $Z_{\mathbf{w}}(\mathbf{x})$ is calculated as:{\color{black}
\begin{equation}\label{eqn:linear_crf}
Z_{\mathbf{w}}(\mathbf{x}) = \sum_{\mathbf{y}}\exp\ \smashoperator{\sum_{(y',y,i)\in\mathcal{E}(\mathbf{x}, \mathbf{y})}}\ \mathbf{w}^T\mathbf{f}_{\mathbf{x}}(y',y,i)
\end{equation}}
%

\vspace{-0.5em}\noindent
where $\mathbf{f}_{\mathbf{x}}(y', y, i)$ represents the feature vector on the edge connecting state $y'$ at position $i-1$ to state $y$ at position $i$. The time complexity of the inference procedure for this model is $O(n\left\vert\mathcal{Y}\right\vert^2)$. 
\subsection{Semi-CRF}

In semi-CRF \cite{Sarawagi2004}, in addition to the edges defined in linear CRF, there are additional edges from a node to all nodes up to $L$ next words away, representing a segment within which the words will be labeled with a single label.

The normalization term $Z_{\mathbf{w}}(\mathbf{x})$ is calculated as:{\color{black}
\begin{equation}\label{eqn:semi_crf}
Z_{\mathbf{w}}(\mathbf{x}) = \sum_{\mathbf{y}\in\mathcal{Y}^n}\exp\ \ \ \ \smashoperator{\sum_{(y',y,i-k,i)\in\mathcal{E}(\mathbf{x},\mathbf{y})}}\ \ \ \ \mathbf{w}^T\mathbf{g}_{\mathbf{x}}(y', y, i-k, i)
\end{equation}}

\vspace{-0.5em}\noindent 
where $\mathbf{g}_{\mathbf{x}}(y', y, i-k, i)$ represents the feature vector on the edge connecting state $y'$ at position $i-k$ to state $y$ at position $i$. The time complexity for this model is $O(nL\left\vert\mathcal{Y}\right\vert^2)$.

\subsection{Weak Semi-CRF}
\label{weak-semi-crf}


Note that in semi-CRF, each node is connected to $L\times \left\vert\mathcal{Y}\right\vert$ next nodes. Intuitively, the model tries to decide the next segment length and type at the same time. We now propose a weaker variant that makes the two decisions separately by restricting each node to connect to either only the nodes of the same label up to $L$ next words away, or to all the nodes only in the next word. We call this variant {\em Weak Semi-CRF}.

To implement this, we need to split the original nodes into Begin and End nodes, representing the start and end of a segment. The End nodes connect only to the very next Begin nodes of any label, while the Begin nodes connect only to the End nodes of same label up to next $L$ words. {\color{black}We denote the set of the earlier edges as $\mathcal{E}_A(\mathbf{x},\mathbf{y})$ and the latter edges as $\mathcal{E}_J(\mathbf{x}, \mathbf{y})$.
The normalization term $Z_{\mathbf{w}}(\mathbf{x})$ is then:
\begin{equation}\label{eqn:wsemi_crf}
\begin{array}{rl}
Z_{\mathbf{w}}(\mathbf{x}) = \displaystyle\sum_{\mathbf{y}\in\mathcal{Y}^n}\exp\!\!\!\!\! & \displaystyle
\biggl(\ \smashoperator[r]{\sum_{(y',y,i)\in\mathcal{E}_A(\mathbf{x},\mathbf{y})}} \mathbf{w}^T\mathbf{f}_{\mathbf{x}}(y', y, i)\biggr. \\
\!\!\!\!\! & \displaystyle\biggl. + \smashoperator[r]{\sum_{(y,i-k,i)\in\mathcal{E}_J(\mathbf{x},\mathbf{y})}} \mathbf{w}^T\mathbf{g}_{\mathbf{x}}(y, i-k, i)\biggr)
\end{array}
\end{equation}}

\vspace{-0.5em}
%
%
%
%
\noindent 
where $\mathbf{g}_{\mathbf{x}}(y, i-k, i)$ represents the feature vector on the edge connecting the Begin node with state $y$ at position $i-k$ to the End node with the same state $y$ at position $i$.
Note that, different from the $\mathbf{g}_\mathbf{x}$ function defined in Equation (\ref{eqn:semi_crf}), this new $\mathbf{g}_\mathbf{x}$ function is defined over a single (current) $y$ label only, making the time complexity $O(n\left\vert\mathcal{Y}\right\vert^2 + nL\left\vert\mathcal{Y}\right\vert)$.
Theoretically this model is slightly more efficient than the conventional semi-CRF model.

Unlike conventional (first-order) semi-Markov CRF, this new model does not allow us to capture the dependencies between one segment and its adjacent segment's label information.
We argue that, however, such dependencies can be less crucial for our task.
We will empirically assess this aspect through experiments.
Figure \ref{fig:diffs} illustrates the differences among the three models.

%
%

\section{Features}
In linear CRF, the baseline feature set considers the previous word, current word, and the tag transition.

In semi-CRF, following \newcite{Sarawagi2004} we put each word which is not part of a noun phrase in its own segment, and put each noun phrase in one segment, possibly spanning over multiple words. 
Here we set $L=6$ and ignored NPs with more than six words during training, which is less than 0.5\% of all NPs.
For each segment, we defined the following features as the baseline: (1) indexed words inside current segment, running from the start and from the end of the segment, (2) the word before and after current segment, and (3) the labels of previous segment and current segment.

In weak semi-CRF we use the same feature set as semi-CRF, adjusting the features accordingly where segment-specific features (1) are defined only in the Begin-End edges, and transition features (3) are defined only in the End-Begin edges.

For each model we then add the character prefixes and suffixes up to length 3 for each word (\verb=+a=), Brown cluster \cite{Brown1992} for current word (\verb=+b=), and word shapes (\verb=+s=). For Brown cluster features we used 100 clusters trained on the whole NUS SMS Corpus. The cluster information is then used directly as a feature.

Word shapes can be considered a generic representation of words that retains only the ``shape'' information, such as whether it starts with capital letter or whether it contains digits. The Brown clusters and word shapes features are applied to each of the word features described in each model.

\section{Experiments}

All models were built by us using Java,
and were optimized with L-BFGS.
Models are all tuned in the development set for optimal $\lambda$. The optimal $\lambda$ values are noted in Table \ref{reg-param}.

Since the models that we consider are all word-based \footnote{We experimented with character-based models, but they do not perform well. We leave them for future investigations.}, we tokenize the corpus using a regex-based tokenizer similar to the \verb=wordpunct_tokenize= function in Python NLTK package. We also included some rules to consider special anonymization tokens in the SMS dataset \cite{Chen2013}.

The gold character spans are converted into word labels in BIO format, reducing or extending the character spans as necessary to the closest word boundaries. 
The converted annotations are regarded as gold word spans. 
Note that this conversion is lossy due to the presence of \textit{improper} NPs, which makes it impossible for the converted format to represent the original gold standard.

We evaluated the models in the original character-level spans and also in the converted word-level spans, to see the impact of the lossy conversion on the scores. In character-level evaluation, the system output is converted back into character boundaries and compared with the original gold standard, while in the word-level evaluation, the system output is compared directly with the gold word spans. 
For this reason, we anticipate that the scores in word-level evaluation will be higher than in the character-level evaluation.
The results are shown in Table \ref{result}. The scores for ``Gold'' in the character-level evaluation mark the upperbound of word-based models due to the presence of \textit{improper} NPs.

The average time per training iteration on the base models is 1.311s, 2.072s, and 1.811s respectively for Linear CRF, Semi-CRF, and Weak Semi-CRF.

\begin{table}
\footnotesize
\begin{tabular}{r|l|l|l}
 & Linear CRF & Semi-CRF & Weak Semi-CRF \\\hline
\verb=base= & 0.125 & 2.0 & 2.0 \\
\verb=    +s= & 0.25 & 1.0 & 2.0 \\
\verb=  +b  = & 0.5 & 1.0 & 2.0 \\
\verb=  +b+s= & 0.5 & 2.0 & 2.0 \\
\verb=+a    = & 1.0 & 2.0 & 2.0 \\
\verb=+a  +s= & 2.0 & 1.0 & 2.0 \\
\verb=+a+b  = & 1.0 & 2.0 & 2.0 \\
\verb=+a+b+s= & 2.0 & 2.0 & 2.0 \\
\end{tabular}
\caption{Tuned regularization parameter $\lambda$ from the set \{0.125, 0.25, 0.5, 1.0, 2.0\} for various feature sets. \texttt{+a}, \texttt{+b}, and \texttt{+s} refer to the affix, Brown cluster, and word shape features respectively.}
\label{reg-param}
\vspace{-3mm}
\end{table}

\begin{table}[ht!]
\footnotesize
\begin{center}
\scriptsize
\begin{tabular}{r|c|c|c||c|c|c}
& \multicolumn{3}{c||}{Character-level Eval.} & \multicolumn{3}{c}{Word-level Eval.} \\
& \textit{Prec} & \textit{Rec} & \textit{F} & \textit{Prec} & \textit{Rec} & \textit{F} \\\hline
\multicolumn{7}{c}{Linear CRF} \\\hline
\verb=base= & 72.29 & 70.13 & 71.19 & 74.04 & 71.93 & 72.97 \\
\verb=    +s= & 72.56 & 70.50 & 71.52 & 74.38 & 72.38 & 73.36 \\
\verb=  +b  = & 72.48 & 71.82 & 72.15 & 74.32 & 73.77 & 74.04 \\
\verb=  +b+s= & 72.90 & 72.10 & 72.50 & 74.70 & 73.99 & 74.34 \\
\verb=+a    = & 72.56 & 72.41 & 72.49 & 74.66 & 74.62 & 74.64 \\
\verb=+a  +s= & 72.65 & 71.93 & 72.29 & 74.69 & 74.07 & 74.38 \\
\verb=+a+b  = & 72.63 & 72.80 & 72.71 & 74.70 & 75.00 & 74.85 \\
\verb=+a+b+s= & 72.63 & 72.74 & 72.68 & 74.77 & 74.99 & 74.88 \\
\hline
\multicolumn{7}{c}{Semi-CRF} \\
\hline
\verb=base= & 74.94 & 73.80 & \textbf{74.37} & 76.50 & 75.45 & 75.97 \\
\verb=    +s= & 75.14 & 73.48 & 74.30 & 76.81 & 75.23 & 76.01\\
\verb=  +b  = & 73.95 & 74.50 & 74.22 & 75.82 & 76.50 & 76.15 \\
\verb=  +b+s= & 73.79 & 74.08 & 73.93 & 75.67 & 76.09 & 75.88 \\
\verb=+a    = & 74.31 & 75.08 & \underline{\textbf{74.69}} & 76.20 & 77.11 & \underline{\textbf{76.65}} \\
\verb=+a  +s= & 74.36 & 74.49 & \textbf{74.42} & 76.32 & 76.57 & \textbf{76.44}\\
\verb=+a+b  = & 74.30 & 74.88 & \textbf{74.58} & 76.20 & 76.92 & \textbf{76.55} \\
\verb=+a+b+s= & 74.24 & 74.93 & \textbf{74.58} & 76.23 & 77.06 & \textbf{76.64} \\
\hline
\multicolumn{7}{c}{Weak Semi-CRF} \\
\hline
\verb=base= & 74.84 & 73.94 & \textbf{74.39} & 76.47 & 75.67 & 76.07 \\
\verb=    +s= & 74.84 & 72.67 & 73.74 & 76.50 & 74.40 & 75.43 \\
\verb=  +b  = & 74.13 & 74.12 & 74.12 & 75.97 & 76.08 & 76.02 \\
\verb=  +b+s= & 74.19 & 74.21 & 74.20 & 76.06 & 76.19 & 76.13 \\
\verb=+a    = & 74.07 & 75.13 & \textbf{74.60} & 76.02 & 77.23 & \textbf{76.62} \\
\verb=+a  +s= & 74.47 & 74.49 & \textbf{74.48} & 76.44 & 76.58 & \textbf{76.51} \\
\verb=+a+b  = & 74.08 & 74.57 & \textbf{74.32} & 76.01 & 76.64 & \textbf{76.32} \\
\verb=+a+b+s= & 74.19 & 74.43 & \textbf{74.31} & 76.15 & 76.52 & \textbf{76.33} \\
\hline
\hline
Gold & 95.96 & 95.81 & 95.88 & 100.0 & 100.0 & 100.0 
\end{tabular}
\end{center}
\vspace{-1mm}
\linespread{0.9}
\caption{Scores on test set (both character-level and word-level evaluation) using optimal $\lambda$. \texttt{+a}, \texttt{+b}, and \texttt{+s} refer to the affix, Brown cluster, and word shape features respectively. Best F1 scores are underlined, and values which are not significantly different in 95\% confidence interval are in bold}
\vspace{-5mm}
\label{result}
\end{table}



\subsection{Discussion}

First, we see that the two variants of semi-CRF models perform better compared to the baseline linear CRF model, showing the benefit of using segment features over only single word features. 

It is also interesting that, while being a weaker version of the semi-CRF, the weak semi-CRF can actually perform in the same level within 95\% confidence interval as the conventional semi-CRF. This shows that some of the dependencies in the conventional semi-CRF do not really contribute to the strength of semi-CRF over standard linear CRF. As noted in Section \ref{weak-semi-crf}, weak semi-CRF makes the decision on the segment type and length separately. This means there is enough information in the local features to decide the segment type and length separately, and so we can remove some combined features while retaining the same performance.

This result, coupled with the fact that the weak semi-CRF requires 12.5\% less time than the conventional semi-CRF (1.811s vs 2.072s), shows the potentials of using this weak semi-CRF as an alternative of the conventional semi-CRF. With more label types (here only two), the difference will be larger, since the weak semi-CRF is linear in number of label types, while conventional semi-CRF is quadratic.

\vspace{-1mm}

\section{Related Work}\vspace{-1mm}

Ritter et al. \shortcite{Ritter2011a} previously showed that off-the-shelf NP-chunker performs worse on informal text. Then they trained a linear-CRF model on additional in-domain data, reducing the error up to 22\%. 
However no results on semi-CRF was given.

Semi-CRF has proven effective in chunking tasks.
Other variants of semi-CRF models also exist.
Nguyen et al. \shortcite{Nguyen2014} explored the use of higher-order dependencies to improve the performance of semi-CRF models on synthetic data and on handwriting recognition. They exploited the sparsity of label sequence in order to make the training efficient.

It is also known that feature selection is an important aspect when trying to use semi-CRF models to improve on the linear CRF. Andrew \shortcite{Andrew2006} reported an error reduction of up to 25\% when using features that are best exploited by semi-CRF.



\vspace{-1mm}

\section{Conclusion and Future Work}\vspace{-1mm}

In this paper we present a new NP-annotated SMS corpus, together with a novel variant of the semi-CRF model, which runs in significantly lower time while maintaining similar accuracy on the NP chunking task on the new dataset. 
Future work includes the application of the weak semi-CRF model to other structured prediction problems, as well as performing investigations on handling other types of informal or noisy texts such as speech transcripts.
We make the code and data available for download at \texttt{http://statnlp.org/research/ie/}.

\section*{Acknowledgements}

We would like to thank Alexander Binder, Jie Yang, Dinh Quang Thinh as well as the 64 undergraduate students who helped us with annotations. We would also like to thank the three anonymous reviewers for their helpful comments.
This work is supported by SUTD grant SRG ISTD 2013 064 and MOE Tier 1 grant SUTDT12015008. We thank Razvan Bunescu for pointing out an error in Equations \ref{eqn:linear_crf}, \ref{eqn:semi_crf}, and \ref{eqn:wsemi_crf} in an earlier version of this paper.

\bibliographystyle{naaclhlt2016}
\bibliography{naaclhlt2016-weak-scrf-revision-2018}

\end{document}